# Using Online Customer Reviews to Classify, Predict, and Learn about Domestic Robot Failures

Shanee Honig[a]* (ORCID 0000-0003-0674-6623), Alon Bartal[b] (ORCID 0000-0002-1206-1754), Yisrael Parmet[a] (ORCID 0000-0002-2071-7338), Tal Oron-Gilad[a] (ORCID 0000-0002-9523-0161)

[a]Ben-Gurion University of the Negev, Be'er Sheva, Israel.

[b]Bar-Ilan University, Ramat-Gan, Israel.

*Corresponding author. Tel.: +972-585-100890; e-mail: shaneeh@post.bgu.ac.il.

**Abstract**

There is a knowledge gap regarding which types of failures robots undergo in domestic settings and how these failures influence customer experience. We classified 10,072 customer reviews of small utilitarian domestic robots on Amazon by the robotic failures described in them, grouping failures into twelve types and three categories (Technical, Interaction, and Service). We identified sources and types of failures previously overlooked in the literature, combining them into an updated failure taxonomy. We analyzed their frequencies and relations to customer star ratings. Results indicate that for utilitarian domestic robots, Technical failures were more detrimental to customer experience than Interaction or Service failures. Issues with Task Completion and Robustness & Resilience were commonly reported and had the most significant negative impact. Future failure-prevention and response strategies should address the robot's technical ability to meet functional goals, operate and maintain structural integrity over time. Usability and interaction design were less detrimental to customer experience, indicating that customers may be more forgiving of failures that impact these aspects for the robots and practical uses examined. Further, we developed a Natural Language Processing model capable of predicting whether a customer review contains content that describes a failure and the type of failure it describes. With this knowledge, designers and researchers of robotic systems can prioritize design and development efforts towards essential issues.

**Data Availability Statement**
The datasets generated during the current study are available from the corresponding author on reasonable request.

**Funding**
No research funding was received for conducting this study. The first author was supported by scholarships as noted in the acknowledgments.

**Acknowledgments**
The first author is supported by The Helmsley Charitable Trust through the Agricultural, Biological, Cognitive Robotics Initiative and the Marcus Endowment Fund, and by Ben-Gurion University through the High-tech, Biotech and Chemo-tech Scholarship.

# Using Online Customer Reviews to Classify, Predict, and Learn about Domestic Robot Failures

1. **Introduction**

Affordable robots for domestic tasks are becoming increasingly common. It is estimated that sales of domestic robots will exceed 55 million units by 2022 [1]. As utilitarian domestic robots grow in popularity, the question of what to do when they fail becomes increasingly relevant. Current fault-diagnosis and failure-handling systems in robots are typically designed for professional users [2]. However, relying on professionals to understand and resolve robots' faulty behaviour in households is not ideal, as it delays users' ability to accomplish their goals and increases customer support costs.

We aim to advance the collective knowledge on how to develop resilient systems that enable everyday untrained users to correct and respond to robotic failures in their homes. As a step towards this goal, the present research aims to understand and systematize the types and sources of robotic failures that users experience in domestic environments and how they influence customers' opinions. With this knowledge, designers of robotic systems will be able to extract and prioritize design and development efforts towards prolific and essential issues.

Prior research has investigated the type of failures robots experienced in urban search and rescue [3], military applications [3], robotic surgical systems [4–6] and robotic soccer competitions [7]. While these studies provide valuable insights, they may not apply to experiences with domestic robots since the robot operators were trained professionals. Their criteria of what constitutes a failure could differ from that of the wider public, who are generally inexperienced with robots [8]. Unexpected or incoherent behaviours can be interpreted as erroneous [9], so prior experience and training may influence the identification of incorrect behaviour. In addition, domestic environments are different from those previously surveyed. A typical home is less structured than a surgical lab but more structured than the great outdoors. It contains many unique obstacles that can impact the failure types experienced. Lastly, the context of the use of a household robot differs greatly from the examples previously surveyed. In domestic settings, users can accomplish their goals through non-robotic means, multi-task, and decide whether to control their robots remotely or in person. These parameters may impact the types of failures they perceive and how they are perceived.

Tolmeijer et al. [10] performed a risk analysis of the probability of failures and their impact on user trust in the robot. Their work serves as a good starting point to understand how different failure types can be prioritized and addressed. However, their analysis relied on limited data collected from robots in non-domestic environments, that were not personally owned by the people who participated. Further evidence is warranted to understand whether their conclusions could be generalized to domestic robots. There is limited research evaluating how different robot failures impact the user experience in naturalistic settings [2]. Some laboratory studies have indicated that the type of failure may influence the degree of user trust in the robot [11–13]. It would be beneficial to evaluate the degree to which failure types relate to the overall customer experience in non-laboratory Human-Robot Interactions (HRIs).

To learn more about the types of robotic failures people have experienced in domestic settings and their impact on customer experience, we analyzed 10,072 online customer reviews of various utilitarian domestic robots. Customer reviews often contain descriptions of the issues people have experienced and indicators of customer sentiment [14]. The failures reported in the reviews were classified using thematic analysis [15] and used to form a taxonomy of domestic robot failure types. Relations between failure types and customer experience were evaluated using star ratings as proxies, since prior studies showed that star rating correlates well with sentiments of the content of the online customer review [16, 17]. In addition, we used our dataset to develop Natural Language Processing (NLP) models aiming to automatically identify reviews that contain a description of a failure and the types of failures it contains.

This research extends previous HRI research in three primary ways. First, we created a failure taxonomy tailored to domestic robots and nonexpert users, based on HRI experiences in which failures had real negative consequences. Existing taxonomies are based primarily on interactions with experts in non-domestic environments or laboratory settings with low ecological validity [2, 18]. Second, we present recommendations regarding the types of failures that should be prioritized, based on an in-depth analysis of the relationship between the frequency of a reported failure, its type, and its influence on the star rating. The practice of using online customer reviews to prioritize engineering design and development efforts was established previously [19, 20]. However, to the best of our knowledge, previous research has not applied this methodology to compare the potential impact of robot failure types. Third, we present NLP models that identify whether a given written review (in free text format) contains a failure description and, if so, the types of failures it describes. NLP models were created before to predict various information embedded in product reviews [21–23]. However, we are unaware of research that used this method to identify robotic failures autonomously. The developed NLP models can be used to understand the dynamic shortcomings of domestic robots by autonomously identifying and monitoring failure events of different levels of priority. The outcomes of this work enable the scientific community to observe and learn about real-world robot failures and allow designers to improve upon their products.

The remainder of the paper is structured as follows. Section 2 presents related background information. Section 3 describes how we created and analyzed the dataset and our NLP models. Section 4 presents the failure types identified from the dataset, sections 5-8 portray the failure frequencies, how different failures impact customer experience, our NLP model performance evaluation, and an updated failure taxonomy, respectively. Finally, Section 9 discusses our findings and recommendations, limitations, suggestions for future studies, and conclusions.

## 2. Background

In this section, we review the existing literature regarding robotic failure taxonomies, online customer reviews, and the use of data mining in HRI.

### a. Existing Failure Taxonomies

To understand what types of robot failures are noted by users in domestic settings, we first looked at the existing classifications of robot failures. Our prior work [2] surveyed and merged existing failure taxonomies. The high-level branches of the taxonomy are presented in Figure 1. This mapping distinguishes primarily between *technical failures* and *interaction failures*. Technical failures are caused by hardware errors or problems in the robot's software system. Interaction failures refer to issues that arise from uncertainties in the interaction with the environment, other agents, and humans.

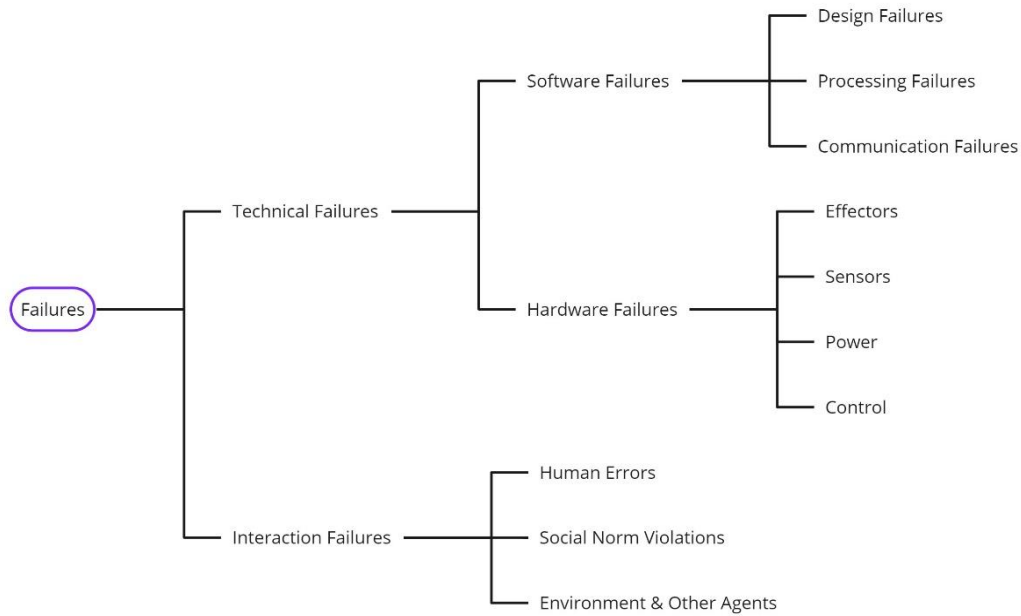

**Figure 1.** A high-level human-robot failure taxonomy after Honig & Oron-Gilad [2].

Since then, additional relevant taxonomies have emerged. Tolmeijer et al. [10] presented a taxonomy of failure types with regards to their impact on trust and trust-related mitigation strategies. They distinguished four different failure types: (1) Design, (2) System, (3) Expectation, and (4) User. *Design failures* are instances in which the system behaves as intended, however, the design choices were not ideal for the HRI. *System failures* are situations in which the system does not act as intended, divided into software and hardware failures. *Expectation failures* refer to situations in which the system works as intended but defies user expectations. These were further divided into omission failures (where the robot does not act when the user expects it) and commission failures (where the robot does something the user does not expect). *User failures* refer to intentional or unintentional instances where the users interact with the system in ways they were not supposed to, either because of a design failure or an expectation failure.

Tian and Oviatt [24] created a taxonomy that focuses on how errors influence social interactions and social relationships in HRI. They identified five main categories: (1) *Breach in empathic and emotional reactions* (e.g., a robot incorrectly recognizing a user's emotions or showing inappropriate emotional expressions during interactions), (2) *Insufficient social skills* (e.g., a robot failing to adjust its behaviour to the social context or exhibiting inappropriate levels of assertiveness), (3) *Misunderstanding the user* (e.g., a robot incorrectly assessing user intentions or failing to respond to joint attention cues), (4) *Insufficient communicative functions* (e.g., a robot producing incomprehensible expressions), and (5) *Breach in collaboration and prosociality* (e.g., a robot violating privacy or failing to maintain reciprocity or fairness in collaboration). These taxonomies will feed into and inform our analysis of the types of failures extracted from online customer reviews.

*b. Online Customer Reviews*

Customer reviews have become part of the online shopping experience and present meaningful opportunities to provide data for research and development. According to surveys by the Pew Research Center and by the European Consumer Centers' Network, 82% of American and European respondents consult online reviews when buying something for the first time [25, 26]. Online reviews contain rich information on purchase experiences, vendors, customer complaints, user experience, and customer satisfaction [27, 28]. Scientific research has analyzed customer reviews to gain insights on product sales [29], product ranking [30], to study customers' opinions and sentiments towards products and services [31–34] and identify product defects [35]. It is not surprising that a growing body of research draws upon online customer review data for insight.

The reviews' reliability is of particular importance to our current investigation. A variety of review biases have been reported [36]. People who think a product is of very low or high quality are more likely to write a review than those who believe the product is of average quality (the 'underreporting' bias; [37, 38]. Highly positive reviews outnumber highly negative reviews, possibly because people who think negatively of a product will not buy it in the first place (the 'purchase bias'; [37]), or because poorly rated products are actively removed. Online customer ratings tend to become more negative over time, as more customers rate the product [39]. Perhaps early reviews come from early adopters, who may be positively biased towards the product [39]. Review fraud may be an alternative explanation, as it becomes more costly to influence the average product ratings as the number of reviews grow [40].

Review fraud is a significant concern in assessing the reliability of online reviews. Although online reviews are meant to be real testaments of others' experiences with a product or service [41], there are many instances in which fake reviews are posted to mislead potential customers [42]. Fake reviews can be positive (to increase a company's profits) or negative (to decrease a competitor's profit). The proportion of fake reviews ranges from 16% to 33% [42]. They are more common for smaller companies, with inferior products and fewer reviews [42]. Despite the potential reliability concerns, companies and researchers use online customer reviews to obtain insights in a wide variety of fields [43–46]. Most reviews are genuine, and various solutions exist for identifying and filtering out potential fake reviews [42, 47–50].

*c. Analysis of Customer Reviews in HRI*

Within HRI, limited research targeted online consumer reviews for identifying and analysing customer experiences. One recent exception is [17] who analyzed online TripAdvisor reviews to determine customers' sentiments on hotels employing robotic services. Their analysis identified feelings towards various service quality attributes (e.g., the robot's design and speed) and operational attributes (such as luggage handling and reception). It revealed that hotel service robots underperformed in their communication and interaction abilities, consistent with the lack of research in resolving interaction failures [2].

Another recent study [20] aimed to explore which robot types and features are most successful in the consumer market. As part of their investigation, they performed sentiment analysis on 13,605 product reviews of domestic robots from two crowdfunding sites (Kickstarter and Indiegogo), to identify product features consumers found necessary. Although robot failures were not their primary focus, they found that words related to product charging and network connection ('battery' / 'connect' / 'charge') co-occurred most frequently with the words 'problem' and 'issue'.

Recently, Carames et al. [51] performed a thematic analysis of 1740 online reviews to understand the types of experiences people have as they interact with robot vacuums. They looked at situations that would require users to troubleshoot and found that 37% can be classified as positive experiences, and 63% as negative. Their analysis showed that users who had positive experiences were willing to change their behavior or surroundings to solve an issue, whereas those who had negative experiences were not. Positive experiences did not necessarily require the troubleshooting solution to be successful, nor did negative experiences necessarily involve a failure to troubleshoot the robot. Another interesting finding was that higher-priced robots had more reviews that mentioned negative troubleshooting experiences than cheaper robots. The authors gave two possible reasons: (1) that higher-priced robots also have higher levels of automation, which make troubleshooting more complicated, and (2) that higher-cost increases user expectations, reducing willingness to alter behavior and troubleshoot.

Prior investigations [17, 20, 51] confirm the potential of extracting meaningful information from online reviews to improve the quality of robotic services. However, none of them focused on identifying or describing specific causes of negative sentiments. Our exploratory research aims to fill this gap.

## 3. Methodology

*a. Dataset Development and Manual Coding*

To learn about the types of robotic failures people have experienced in domestic settings and their impact on customer experience, we downloaded ~15,000 online customer reviews from Amazon.com between November 2018-April 2020. Using a python script, we harvested reviews of four domestic robot types: vacuum cleaners, pool cleaners, lawnmowers, and grill cleaners. Amazon was selected because of its size and popularity and because it is a commonly used source for analyzing customer experiences [36, 45]. We chose the robot types and specific models based on the number of reviews on Amazon at the time of collection. For each review, we retrieved the buyer's username, posting date, star rating, review title and text. Reviews not in English, duplicate reviews, and reviews that described a different product were manually removed from the dataset (< 2%). Identifying information was then stripped from the datasets, leaving only the star rating, posting date, and review content.

We manually examined 10,072 customer reviews (Table 1), noting instances of robot failures described in them (if any). Coded failures included system behaviors, characteristics, and services perceived by customers to differ from the ideal, normal, or correct functionality. This is consistent with the definition of failure by [52]. Only affirmed instances of specific failures were coded. We coded failures based on their perceived importance, labelling them differently if a failure was described as part of a complaint (marked by '1') or a comment (marked by '2'). A complaint is when the reviewer was not satisfied with an issue, e.g., *"I hate that the robot doesn't return to its charger."* A comment is when the reviewer mentioned the case as a side-note or said something to indicate it was not critical, e.g., "*Keep in mind you have to be home when you run the robot because it doesn't always return to the charger,*" or "*It's a bit too heavy for my wife, but I can take the robot out of the pool easily, so we don't mind.*".

**Table 1.** Dataset Description

| Review Count | Robot Types | Robot Models |
|---|---|---|
| **8306** | Vacuum Cleaner | Ecovacs Deebot N79S, iRobot Roomba 690, iRobot Roomba i7+, eufy BoostIQ RoboVac, ILIFE V3s, Amarey A800 Robot Vacuum, iRobot Roomba 614, iRobot Roomba 675 |
| **1357** | Pool cleaner | Aquabot AJET122 Pool Rover S2, Dolphin Premier Robotic Pool Cleaner, Polaris F9550 Sport, Solar Breeze |
| **329** | Grill Cleaner | Grillbot Automatic Grill Cleaning Robot |
| **80** | Lawn Mower | Worx WR140, WR150, WR153 Landroid M 20V Power Share Robotic Lawn Mower, Husqvarna 450X, Husqvarna AUTOMOWER 326, Husqvarna AUTOMOWER 335 |

*b. Failure Classification*

Once the dataset was complete, each failure was labelled by two independent annotators into categories and types. A third person resolved discrepancies.

**Failure categories** were initially differentiated between *interaction failures* and *technical failures* based on Honig & Oron-Gilad [2]. A third category, *service failures*, was added during the labeling process due to a need to address problems customers had related to the service provider's decisions, actions, or lack of actions. We initially attempted to classify failures by the high-level categorization suggested by Tolmeijer et al. [10], differentiating between design, system, expectation, and user failures. However, we found that it is often difficult to distinguish these categories based on the information in online reviews since customers only report on the part of the relevant information necessary to identify the source of the problem.

**Failure types** were identified using thematic analysis [15]. The goal of this procedure was to gain insight into the problems people described in their reviews through exploratory means. We did not want to limit insights by relying solely on existing taxonomies. Failure type classifications (e.g., Task Completion) were performed separately from failure category classifications (e.g., Technical, Interaction, or Service) so that one process doesn't influence the other. The source of the problem cannot be definitively confirmed from reviews, so multiple classifications were possible. We attributed failures to all possible categories and types that may explain the source of the matter.

Some of the problems that were included under *service failures* are not the direct source of robot failures; instead, they are general issues that people have with the service provider, such as issues with the robot's pricing and the items included in the robot's package. The analysis included these issues because they may indirectly lead to failures by altering user behaviors and expectations. For example, pricing perceptions may influence the likelihood of users adhering to recommended maintenance routines, which could prevent or promote a wide variety of failures. Pricing and service quality have impacted product usage in the past [53, 54]. Similarly, if vacuum cleaning robots that rely on external sensors to define cleaning boundaries are sold to consumers without enough units to roam the operating space safely, it could lead to failure (e.g., to the robot falling down a staircase).

*c. NLP Model Development and Evaluation*

The vacuum cleaner dataset (8,306 reviews, see Table 1) was used to develop NLP models that aimed to identify whether a review contains a description of a failure or not, and the type of failure it describes. Reviews of other robots from the database were not included in the model development process due to their relatively small numbers.

The NLP models were developed using BERT [55] – a deeply bidirectional neural network pre-trained on a large body of unlabeled text, classifying text from relatively small sample sets. We fine-tuned the case-sensitive BERT base-cased model on a downstream binary classification task and created a model to predict whether a review contains a failure or not (i.e., a binary decision of failure or no failure). We also created three models to determine: whether a given review has 1) a technical failure (yes or no), 2) an interaction failure (yes or no), and 3) a service failure (yes or no), respectively. In cases with relatively fewer examples of the target failure category (i.e., Interaction and Service failures), the data used for model development was balanced to contain an equal number of reviews with and without the target failure category. Reviews without the target failure category were randomly selected.

Before creating the models, the data were divided into training (70-80%) and testing (20-30%) sets. The size of the training set was determined by the number of target examples we had for each model. In all cases, 90% of the training set was used to fine-tune the model and 10% was used to validate its performance. For fine-tuning, we used four epochs with a batch size of 32, a sequence length of 128, and a learning rate of $3e^{-5}$. The testing set was used to evaluate the model's predictive performance on new reviews the model had not yet encountered. NLP model evaluation was done in terms of Accuracy (ratio of correctly predicted observation to the total observations), Precision (ratio of correctly predicted positive observations to the total predicted positive observations), Recall (ratio of correctly predicted positive observations to all observations), and F1 score (a weighted average of Precision and Recall).

*d. Analyzing Relations between Failure Classifications and Customer Experience*

Once the dataset was complete, we sought the relationship between failure classifications (failure category and type) and customer experience. Since prior studies had found that star ratings correlate well with sentiments of the full content of the online customer reviews [16, 17], star ratings served as a proxy for customer experience. Failures coded as comments (rather than complaints) were excluded from the analysis to focus on reviewers' negative experiences.

To evaluate whether our classifications can predict star rating, we used three statistical models: Logistic regression, Random Forest (RF), and Support Vector Machine (SVM). For logistic regression, we fitted an ordinal logistic regression and a multinomial logistic regression. For Random Forest and SVM, we fitted classification and regression. In both RF models, we performed an initial search of the hyper parameters on the training set using 10-fold cross validation sampling method. Before fitting the models, the data were divided into training (80%) and validation (20%) sets. Learning was done on the training set, and validation was done on each set separately. Validation was done using several indices: Cohen's Kappa, weighted Cohen's Kappa, Cramer's V, total accuracy rate, and the mean accuracy and sensativity rates of each star level (precision level for each star level) and their positive predicted value (PPV).

A Spearman's rank-order correlation test was applied to determine the relationship between star rating and the number of complaints and comments per review. A Wilcoxon signed rank test was used to compare the median number of complaints per review with the median number of comments per review. Independent samples median tests were used to compare the number of failures across different robot types.

## 4. Failure Types identified from Online Reviews

From the online reviews, we identified 119 recurring failures, which we grouped into 12 failure types. Types and examples are detailed in Table 2. For a breakdown of the 119 failures by type, category, and robot, see Appendix.

**Table 2** The Failure Types derived from the Online Customer Reviews

| Failure Type | Description | Examples |
| --- | --- | --- |
| Task completion | The robot experienced issues completing the task it was designed to accomplish. | The robot didn't clean/mow everywhere; the vacuum robot spreads dirt, gets stuck, doesn't return to the docking station. |
| Robustness & Resilience | Issues related to the robot's ability to tolerate perturbations or maintain operation and structural integrity over time. | The robot no longer turns on, parts breaking or falling off, specific functions no longer work, inconsistent performance, frequent errors, scratches easily, buttons getting stuck. |
| User effort | Using the robot requires substantial effort from the user. | Hard to clean or relocate the robot, require too much supervision, user needs to do manual cleaning or mowing after the robot is done, the button is hard to press, it takes more time to clean the robot's brushes than it would to vacuum manually. |
| Incompatibility | Not compatible with the operating environment or other agents in it. | The robot is too loud, activates the security alarm, scares pets, cannot operate on stairs, cannot clean certain types of pools or floors. |
| Design | Design choices are not ideal for customers. The robot or its App is missing a specific function the user wants. The user would like the robot or App to be designed differently. | Robot doesn't have a shutdown button, the dirt collection bucket is too small, the battery isn't strong enough to clean the entire home on a single charge, didn't like the robot's colour, App isn't supported by the users' devices. |
| Task completion method | Issues with the way the robot completes the task. | Robot moves randomly or unexpectedly in the space, doesn't learn the space, takes too long to complete a task, invests too little time, continuously goes to problematic areas. |
| Market value | Issues related to the pricing and how the robot compares to alternative solutions. | Price doesn't match the perceived value of the robot, and replacement parts are expensive, another robot or model performs better or offers more for the price point, a robotic vacuum is not as strong as a regular vacuum. |
| Safety | Robot gets into situations that may hurt itself, someone else or lead to damage. Including both realized and potential events. | Ran into walls and obstacles, moved or "ate" items in its surrounding, got caught on fire, fell down the stairs, damaged floors, caused physical pain to a person or animal, or scared them. |
| Charging and Battery | Issues relating to the robot's charger or battery. | Battery drains quickly, the robot won't charge, problems connecting the robot to a charger. |

| Failure Type | Description | Examples |
|---|---|---|
| User confusion | Issues that lead to user misunderstandings of the robot or the situation. | Incompatibility between the robot feedback and its behavior, unclear indicators, the robot is hard to learn, and doesn't come with a manual for operating or fixing. |
| Service quality | Issues with the quality of the service the robotic company provides. | Issues with customer support, robot arrived in a used condition, privacy and security concerns, delivered broken or with loose parts or without all items, lacking personalization. |
| Connection & Updates | Issues with connecting the robot to the network, the App, or other compatible devices. | Problems connecting to Google voice control / Alexa / dedicated App to control the robot, fails to connect to Wifi or disconnects from Wifi, App doesn't connect to Robot, infrequent updates, and upgrades, issues with completing the software update. |

Correlations between failure categories (Technical, Interaction, and Service) and failure types (12 types as noted in Table 2) are shown in Figure 2. Service failures were mostly associated with Design Issues, Service Quality, and Market Value. Technical failures were most related to Task Completion, and Robustness & Resilience. Interaction failures were related to Task Completion Method and User Effort.

| | Technical | Interaction | Service |
|---|---|---|---|
| Task completion | 0.78 | 0.44 | 0.32 |
| Robustness & Resilience | 0.53 | 0.17 | 0.31 |
| User effort | 0.33 | 0.61 | 0.25 |
| Incompatibility | 0.33 | 0.43 | 0.19 |
| Design | 0.29 | 0.28 | 0.73 |
| Task completion method | 0.38 | 0.68 | 0.18 |
| Market value | 0.27 | 0.2 | 0.67 |
| Safety | 0.44 | 0.28 | 0.12 |
| Charging and Battery | 0.31 | 0.17 | 0.11 |
| User confusion | 0.28 | 0.45 | 0.19 |
| Service quality | 0.16 | 0.08 | 0.49 |
| Connection & Updates | 0.28 | 0.18 | 0.07 |

**Figure 2** Pearson correlations of failures in different types and categories. Darker cells indicate higher correlations.

## 5. Failure Frequencies by Star Rating, Robot Type, Failure Type, Failure Category, and Perceived Importance

The number of failures by star rating and robot type is described in Table 3. Most of the dataset (65%) contained 5-star reviews, consistent with most Amazon products. Not surprisingly, a negative correlation was found between star ratings and the number of failures ($r_s = -.625$, $p = .000$), complaints ($r_s = -.705$, $p = .000$), and comments ($r_s = .49$, $p = .000$) per review. Still, 24% of 5-star reviews contained descriptions of failures, and 6% of 1-star reviews did not. The median number of complaints per review was significantly higher than comments per review ($Z=-30.1$, $p=0.00$).

**Table 3:** Failure Dataset by Stars and Robot Type

| Star Rating / Robot Type | All Reviews (% of total) | Reviews with Failures (% per star rating) | Mean (SD) Failures per Review | Mean (SD) Complaints per Review | Mean (SD) Comments per Review |
|---|---|---|---|---|---|
| **All** | **10,072** | **4,549 (45%)** | **0.96 (1.53)** | **0.72 (1.43)** | **0.24 (0.62)** |
| 1 | 1,051 (10%) | 988 (94%) | 2.37 (1.93) | 2.34 (1.93) | 0.03 (0.23) |
| 2 | 562 (6%) | 543 (97%) | 2.71 (1.98) | 2.63 (1.98) | 0.07 (0.37) |
| 3 | 568 (6%) | 521 (92%) | 2.26 (1.92) | 2.07 (1.92) | 0.19 (0.56) |
| 4 | 1360 (14%) | 925 (68%) | 1.38 (1.54) | 0.85 (1.33) | 0.52 (0.87) |
| 5 | 6531 (65%) | 1,572 (24%) | 0.39 (0.86) | 0.16 (0.57) | 0.23 (0.59) |
| Vacuum Cleaner | 8,306 (82%) | 3,543 (43%) | 0.85 (1.40) | 0.62 (1.30) | 0.23 (0.59) |
| Pool cleaner | 1,357 (13%) | 769 (57%) | 1.49 (1.87) | 1.22 (1.84) | 0.27 (0.69) |
| Grill Cleaner | 329 (3%) | 169 (51%) | 1.04 (1.45) | 0.85 (1.34) | 0.19 (0.51) |
| Lawn Mower | 80 (0.8%) | 68 (85%) | 3.11 (3.06) | 2.05 (3.15) | 1.06 (1.45) |

Although most reviews were of robotic vacuum cleaners, we found differences in the median number of failures ($\chi^2$=150, df=3, p=0.00), complaints ($\chi^2$=145, df=3, p=0.00), and comments ($\chi^2$=57, df=3, p=0.00) per review among the four types of robots. Pairwise comparisons revealed that vacuum cleaners differed from the other robot types in the median number of complaints per review, and lawnmowers differed from the other robot types in the median number of comments per review (Table 4).

**Table 4** Statistical comparisons of the number of failures per review across different robots (* = $p < .01$, df=1)

| | Independent Samples Median Test - $\chi^2$ | | |
|---|---|---|---|
| **Comparison** | **Failures** | **Complaints** | **Comments** |
| Vacuum Cleaner – Pool Cleaner | 92.7 * | 105.83 * | 1.37 |
| Vacuum Cleaner – Lawn Mower | 57.9 * | 26.98 * | 54.16 * |
| Vacuum Cleaner – Grill Cleaner | 9.8 * | 26.62 * | 1.91 |
| Pool Cleaner – Grill Cleaner | 15.1 * | 0.05 | 3.24 |
| Pool Cleaner – Lawn Mower | 27.7 * | 4.94 | 42.38 * |
| Grill Cleaner – Lawn Mower | 48.0 * | 4.65 | 45.5 * |

The number of failures by robot category, robot type, and failure type is summarized in Figure 3. Technical failures were the most common, both in terms of their relative percentage (49%) and in terms of the rate of reviews that contained them (of all reviews that described failures, 74% described technical failures, 49% described interaction failures, and 37% described service failures). The most common failure types were Task completion, Robustness & Resilience, and User effort.

The vast majority of failures were presented as complaints rather than comments. The distribution between complaints and comments was not equal across classifications. Interaction failures had a lower percentage of complaints than Technical and Service failures. Service quality, Market value, and Robustness and Resilience had the highest percentage of complaints, whereas User Effort, Incompatibility, and Safety had the highest comments rate.

There were many similarities among robots in terms of the failures their users experienced. For example, for all robots, technical failures were the most common. There were also a few differences. The lawnmower robots had relatively more interaction failures and fewer technical failures than the other robots. In terms of failure types, lawnmowers had relatively more User Effort failures and fewer Task Completion failures than the others. The pool cleaners and grill cleaners had more failures related to Robustness & Resilience and fewer failures related to Task Completion Method than the vacuum cleaners and lawnmowers. The vacuum cleaners and grill cleaners had relatively more Incompatibility failures than the pool cleaners and lawnmowers.

| Failure Category and Type | All Failures | | Complaints | | | Comments | | |
|---|---|---|---|---|---|---|---|---|
| | Count | % of Total | Count | % of All Categories | % of All Failures | Count | % of All Categories | % of All Failures |
| Technical | 5083 | 49% | 3940 | 51% | 78% | 1143 | 45% | 22% |
| Service | 2134 | 21% | 1737 | 22% | 81% | 397 | 16% | 19% |
| Interaction | 3073 | 30% | 2057 | 27% | 67% | 1016 | 40% | 33% |
| Task Completion | 2416 | 23% | 1816 | 23% | 75% | 600 | 23% | 25% |
| Robustness & Resilience | 1503 | 14% | 1272 | 16% | 85% | 231 | 9% | 15% |
| User Effort | 1223 | 12% | 697 | 9% | 57% | 526 | 20% | 43% |
| Design Issues | 929 | 9% | 718 | 9% | 77% | 211 | 8% | 23% |
| Incompatibility | 912 | 9% | 574 | 7% | 63% | 338 | 13% | 37% |
| Task Completion Method | 823 | 8% | 621 | 8% | 75% | 202 | 8% | 25% |
| Market Value | 606 | 6% | 517 | 7% | 85% | 89 | 3% | 15% |
| Safety | 554 | 5% | 373 | 5% | 67% | 181 | 7% | 33% |
| Charging & Battery | 394 | 4% | 330 | 4% | 84% | 64 | 2% | 16% |
| User Confusion | 418 | 4% | 338 | 4% | 81% | 80 | 3% | 19% |
| Service Quality | 382 | 4% | 346 | 4% | 91% | 36 | 1% | 9% |
| Connection Issues & Updates | 260 | 2% | 206 | 3% | 79% | 54 | 2% | 21% |

| Failure Category and Type | All Robots | | Vacuum Cleaner | | Pool Cleaner | | Lawn Mower | | Grill Cleaner | |
|---|---|---|---|---|---|---|---|---|---|---|
| | Failures | % of All Failures | Failures | % of All Failures | Failures | % of All Failures | Failures | % of All Failures | Failures | % of All Failures |
| Technical | 5083 | 49% | 3786 | 50% | 1030 | 50% | 91 | 35% | 176 | 49% |
| Service | 2134 | 21% | 1387 | 18% | 598 | 29% | 65 | 25% | 84 | 24% |
| Interaction | 3073 | 30% | 2423 | 32% | 447 | 22% | 106 | 40% | 97 | 27% |
| Task Completion | 2416 | 23% | 1819 | 24% | 502 | 22% | 30 | 12% | 65 | 18% |
| Robustness & Resilience | 1503 | 14% | 876 | 12% | 513 | 23% | 29 | 11% | 85 | 23% |
| User Effort | 1223 | 12% | 835 | 11% | 290 | 13% | 59 | 23% | 39 | 11% |
| Design Issues | 929 | 9% | 653 | 9% | 221 | 10% | 26 | 10% | 29 | 8% |
| Incompatibility | 912 | 9% | 749 | 10% | 111 | 5% | 11 | 4% | 41 | 11% |
| Task Completion Method | 823 | 8% | 712 | 9% | 85 | 4% | 20 | 8% | 6 | 2% |
| Market Value | 606 | 6% | 328 | 4% | 235 | 10% | 20 | 8% | 23 | 6% |
| Safety | 554 | 5% | 486 | 6% | 12 | 1% | 22 | 8% | 34 | 9% |
| Charging & Battery | 394 | 4% | 351 | 5% | 27 | 1% | 6 | 2% | 10 | 3% |
| User Confusion | 418 | 4% | 281 | 4% | 107 | 5% | 16 | 6% | 14 | 4% |
| Service Quality | 382 | 4% | 208 | 3% | 135 | 6% | 15 | 6% | 24 | 6% |
| Connection Issues & Updates | 260 | 2% | 249 | 3% | 4 | 0% | 6 | 2% | 1 | 0% |

**Figure 3** Comparison of Failure count and percentage by failure importance (top) and robot type (bottom)

## 6. Impact of Failure Categories and Type on Star Rating

The Random Forest (RF) models produced the best predictive results (Table 5). Failure categories and failure types had a reasonable predictive value of star rating (~70%). The model was better at predicting extreme ratings (high and low) than mid-level ratings, and tended to misclassify 4-stars as 5-stars, and 2-stars as 1-stars.

The importance of failure categories and types to star rating predictions and their correlations with star ratings are presented in Figure 4. As expected, all failure classifications are negatively correlated with star ratings. Technical failures had a more negative impact on star rating than service failures and interaction failures. Robustness & Resilience failures negatively impacted star rating more than all other failure types. Failure types with the most negligible impact on star rating included Safety, Connection Issues & Updates, and User confusion, which is expected given their low representation in the dataset. Service failures such as market value, design, and service quality negatively impact star rating more than interaction failures, even though interaction failures were more common in the dataset.

**Table 5** Predicting star rating using failure categories and failure types. Measures obtained from validation sets of the RF Models.

| RF Model metrics<br><br>Failure | Overall Accuracy | Cohen's Kappa | Weighted Cohen's Kappa | Cramer's V | Mean of PPV | Mean of Sensitivity |
|---|---|---|---|---|---|---|
| by Categories (3) | 0.726 | 0.41 | 0.698 | 0.384 | 0.583 | 0.812 |
| by Types (12) | 0.703 | 0.40 | 0.659 | 0.367 | 0.375 | 0.834 |

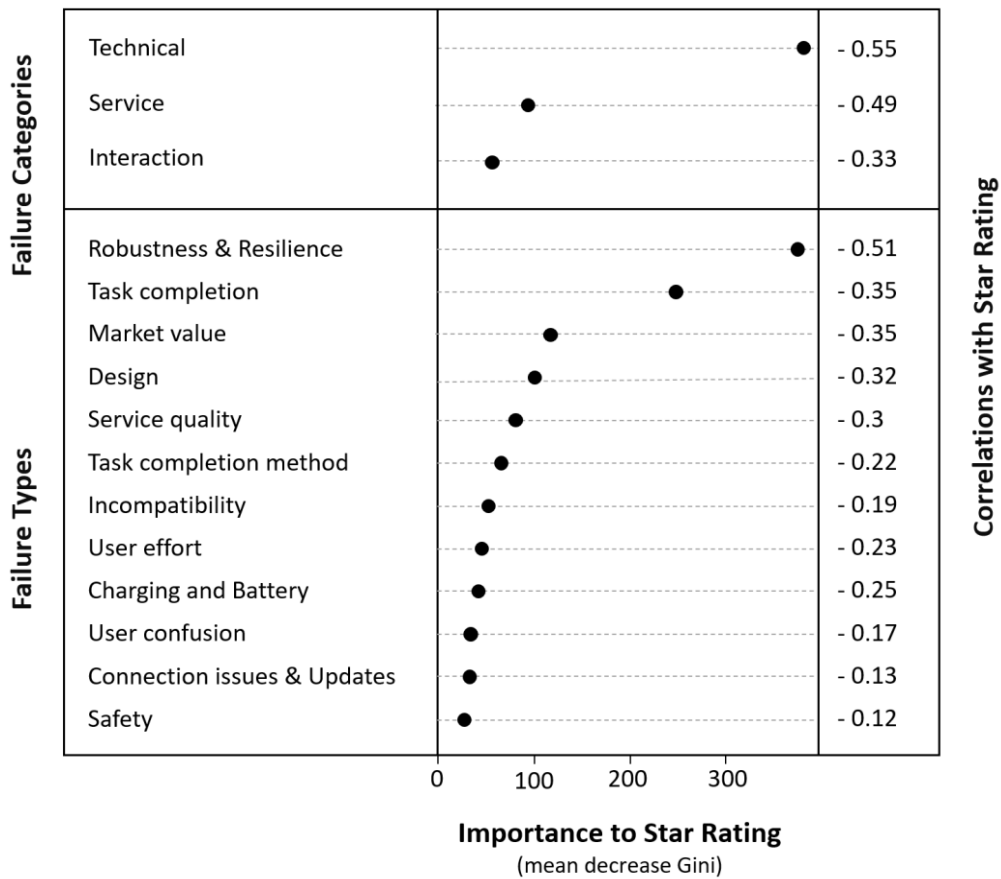

**Figure 4** Variable importance to Star Rating as measured by the Random Forest model, by failure category (top) and failure type (bottom).

## 7. Predicting Presence of Failures using Review Text

Table 6 summarizes the performance of the NLP models. Overall, the final model aimed to predict whether a review contains a failure or not performs quite well, with an F1-score of 0.9. Technical failures were predicted with higher accuracy and precision than interaction and service failures. That said, all three models performed much better than chance. Our code and the resulting models have been uploaded to GitHub (https://github.com/bartala/Domestic_Robot_Failures.git) to be used and improved by the community.

**Table 6** Performance of NLP models created to predict whether a review contains a failure ("Failure Y/N") and to predict whether a review contains a failure of a specific category (Technical, Interaction, Service)

| Predicting | Training Set | Testing Set | Precision | Recall | F1-score | Accuracy |
|---|---|---|---|---|---|---|
| Failure Y/N | 5814 | 2492 | 91% | 90% | 0.90 | 89% |
| Technical | 5814 | 2492 | 79% | 77% | 0.78 | 87% |
| Interaction | 5814 | 2492 | 77% | 80% | 0.78 | 78% |
| Service | 5814 | 2492 | 77% | 78% | 0.78 | 77% |

## 8. Updating the Failure Taxonomies

Analyzing online customer reviews provided a good source of information regarding what types of problems people have experienced with their domestic robots. While it is difficult to understand *why* a particular problem happened, three sources of failure emerged from the online reviews: (1) the *Service Provider*, (2) *Users*, and (3) the *Robot*. Each source has a unique taxonomy of failures, described below. These taxonomies were developed by combining insights from the current investigation with aforementioned theoretical taxonomy developments [2, 10, 24].

*a. Service Provider*

The service providers' roles in robot failures are extensive and can be divided into six subdomains (Figure 5), described below. Sutcliffe and Rugg [56] gave ten environmental and social factors that may increase the likelihood of organizational errors and classified them into *group-level judgment*, *working environment*, and *organizational flaws*. These factors were included previously in our taxonomy under "Interaction failures". They were removed from the current taxonomy since we could not extract examples of how they led to robot failures from the online reviews. For more information on the causes of organizational errors, see Sutcliffe and Rugg [56].

(1) **Robot Design:** failures caused by design choices that were not optimal for HRI.

Table 7 describes the types of design problems that we identified from the online reviews.

**Table 7:** the types of design problems identified from the online reviews.

| Design Consideration | Description | Examples |
|---|---|---|
| Privacy & Security | The design of the system doesn't adequately meet privacy and security needs. | Company sharing traceable or identifiable data about the users' household. Lack of security measures to prevent the robot from being hacked or stolen (in the case of outdoor robots). |
| Maintenance & Storage Requirements | Requirements are incompatible with users' homes or abilities. | Pool cleaning robot must be stored in a shaded place away from water; brushes have to be cleaned multiple times during a cycle, hard to take out parts that require regular cleaning. |
| Repairability | The robot cannot be easily repaired due to robot design decisions. | The robot requires specialized tools to be opened that are not included in the robot package; hardware is glued together so individual components cannot be easily replaced. |
| Robustness & Resilience | The robot does not tolerate perturbations or maintain functional operation and structural integrity over time. | Robot's casing is made from a material that dents or cracks easily, plastic mesh designed to keep filter bags from getting caught in the motor is too large so filter bag gets caught and torn in the motors, wheel housings of pool cleaning robot is made from material that breaks upon impact with pool edge, grill cleaning robot brush bristles are not sufficiently secured in place so they fall off as the robot moves across the grill. |
| Hardware and Software | Hardware and software decisions prevent the robot from being able to operate or implement desired behavior | Vacuum cleaner's navigation algorithm does not guide the robot through all regions in a given space, vacuum cleaner suction power is not sufficient to pick up particles above a certain weight, the robot only being compatible with 5GHz Wifi. |
| Functionality | Mismatch between what the robot is designed to do and user needs / goals. | Robot is not designed to clean stairs, users cannot manually control where the robot works, users cannot schedule the robot to work at different hours during different days. |
| Ergonomics & Usability | The robot is not adapted to the abilities, characteristics, and limitations of its users. | A pool cleaning robot being too heavy to pick up from the pool, operating instructions are not clear or not accessible, the robot is not designed to verbally explain error messages, the robot takes too long to clean a room, requires the app for users to be able to operate critical functions |
| Safety | The robots' safety mechanisms are insufficient, leading to situations where the robot may hurt itself, someone else or lead to damage in the environment. | Robot doesn't recognize stairs causing it to fall down them, the robot wheels are not designed to stop moving when the robot is picked up, grill cleaning robot does not alert when grill is too hot for the robot causing it to melt or catch fire on the grill |
| Aesthetics | The aesthetics of the robot and /or its supporting systems do not match user preferences | Don't like the colour of the robot's casing, the charger and/or robot is too ugly to be placed in public spaces. |
| Physical environment | Issues that result from the robot's design not taking into consideration all aspects of the robot's operating environment, such as changes in… (see below) | |
| | Lighting | Robot is unable to clean rug or floor when it is dark |

|  | Obstacles | Robot "eats" chargers because it wasn't designed to recognize or avoid them |
| --- | --- | --- |
|  | Animals | The robot's charging station can be easily moved by a pet, causing robot to fail to return to it |
|  | Temperature | Grill cleaning robot is unable to clean the grill shortly after it's been used since it wasn't designed to withstand heat |
|  | Weather | Lawn mower does not work well on wet grass |
|  | Terrain | Vacuum cleaner having difficulties picking up dirt from certain types of floors, lawn mower wheels are not secure enough in place for bumpy outdoor environments and continuously fall out |
|  | Type / size of space | vacuum cleaner getting stuck in small spaces such as corridors, or being unable to clean large living room on single charger |
|  | Acoustics | User not hearing auditory alerts of lawn mower due to outdoor noises combined with the noise of the lawn mower itself. |
|  | External systems, including systems designed to integrate with the robot (e.g. Amazon Alexa), systems that share the same operating environment (e.g. other robots that the user operates in the same space), and the infrastructure required to operate the robot (e.g. network connection or electricity). | The robot triggering the user's security alarm, a floor mopping robot washing the floors right before the vacuum cleaner starts to vacuum, unexpected power surges that destroy the robot's charger or battery. |
| Social Environment | Issues that result from the robot's design not accounting for social considerations | Vacuum cleaning robot waking people up with loud error messages in the middle of the night, pool cleaning robot disrupting neighbors with loud auditory alerts. |

(2) **Repair Services:** failures caused by the quality of the repair services made available to customers. When the repair services are not well defined, quick, or accessible, it can create new sets of problems that could have been avoided. Similarly, lack of availability of replacement parts and tools can lead to failure escalation. For example, a customer with a robotic lawn mower had its wheel fall off. The service provider took a long time to send a technician due to a lack of wheels inventory, which led the customer to continue using the robot in the meantime. However, with three wheels, the mower blade began to touch the ground occasionally, which caused it to break and stop working.

(3) **Delivery:** failures caused by the infrastructure used by the service provider to send customers the robot, repair parts, related tools or accessories. Many technical failures can occur if the delivery service doesn't handle items with the necessary care during travel. For example, one of the robot's motors could break if the package accidentally drops during delivery, or the battery could disconnect from the robot if it wasn't secured properly. Delivering the wrong product to the customer or a partial product could also be the source of failure; for example, if the customer was accidentally provided with the wrong replacement battery or received the robot without a filter.

(4) **Quality Control**: errors introduced during production that the service provider did not catch before product deployment. For example, the robot shipped with a faulty sensor or loose parts.

(5) **Communications with Customers**: failures caused by the user receiving insufficient or incorrect information regarding the robot operation from the service provider leading to user mistakes. Information can be communicated (or not) by the service provider via customer service agents, online webpages, a dedicated App, or as part of the unboxing experience (included in or on the packaging, e.g., instruction manuals). For example, some customers stated that their service provider gave incorrect instructions on how to connect their robot to their Google Home device. Some written manuals described features that were not available in the purchased model, leading customers to search for them unsuccessfully.

(6) **Supporting Systems**: failures caused by issues with the robots' supporting systems. Domestic robots rarely arrive as a stand-alone product. Typically, they are accompanied by other technological systems that the service provider designs to integrate with the robot and facilitate its regular operation, such as a charging station, an app or remote to help control the robot, and other technological accessories (e.g. virtual barriers used to define where robotic vacuums are allowed to roam). Service provider failures (design, repair services, delivery, quality control, communications with customers) can also apply to these supporting systems. For example, some reviewers reported having a bug in the robot's App that caused it to change the vacuum cleaning schedule repeatedly, making the robot operate in undesired hours (e.g. waking people up at night) or getting stuck in obstacles (that would have otherwise been cleared before operation).

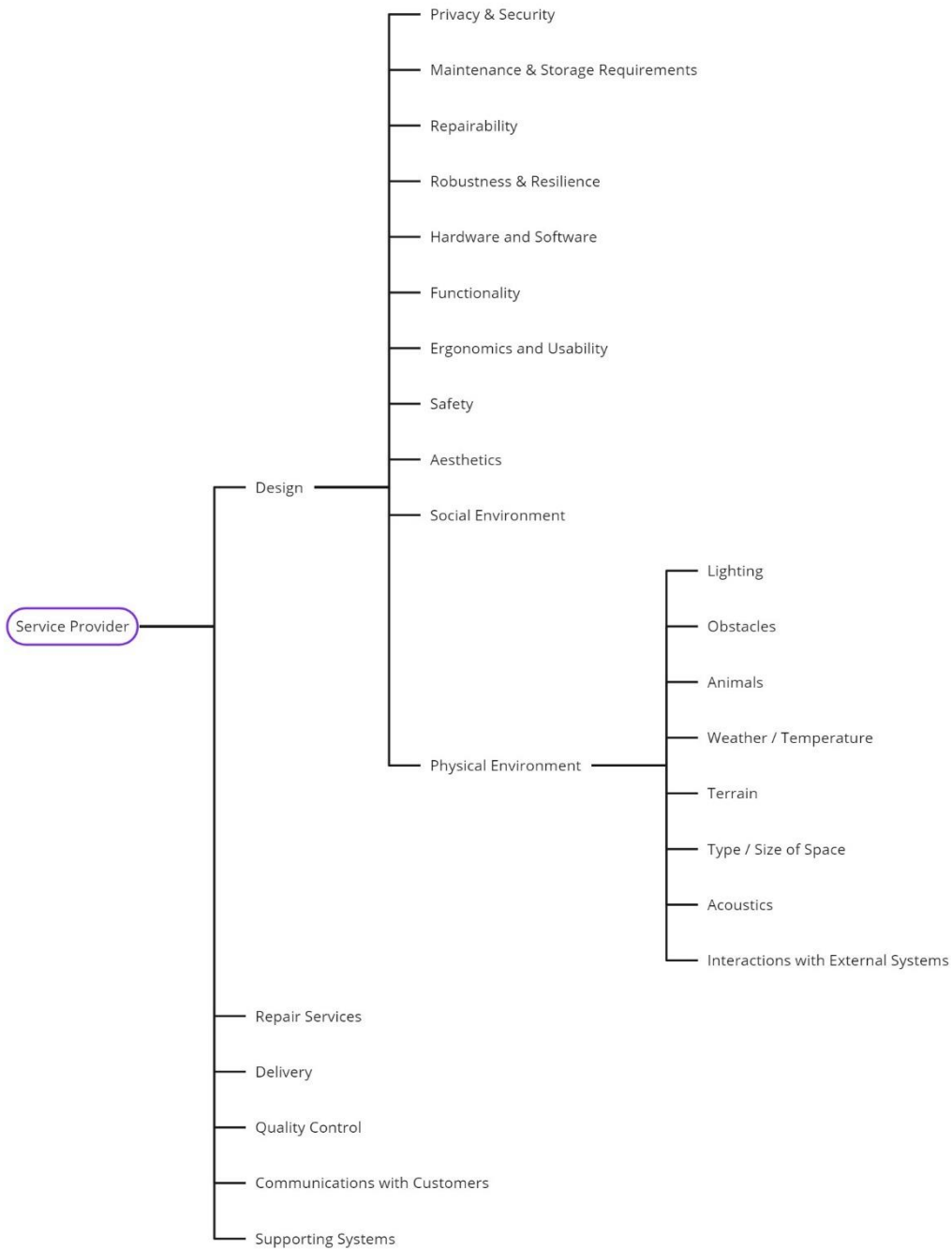

**Figure 5** Sources of Failures caused by the Service Provider.

*b. User*

A third potential source of failure are the users, that is, the people who interact with the robot. Users can be primary users (i.e. people who interact with and/or operate the robot regularly, such as the robot's owner and robot company employees), occasional users (i.e. people who interact with and/or operate the robot occasionally, like the owner's extended family members), or bystanders (people who may have one-time encounters with the robot, e.g. visitors to the home). Similar to [10], user failures are parsed into intentional and unintentional failures (Figure 6). Intentional failures include Violations (deliberate illegitimate actions; e.g., directing the robot to run into a wall) or Lack of Action (purposefully not doing something that needs to be done; prevalent with maintenance routines, e.g., knowing the robot's brushes should be replaced every so often, but deciding not to replace them anyways). Unintentional failures (errors introduced by unintended violations of operating procedures), previously referred to as "Human Errors", include Mistakes (performing a wrong action), Slips (attempting to do the right thing unsuccessfully, e.g., accidentally pressing the wrong button), Lapses (occur as a result of lapses of memory and/or attention; e.g., forgetting to empty the robot's bin), and Expectation Failures (situation in which the system acts as intended, but defies the user's expectation; as detailed in Tolmeijer et al. [10], and Sutcliffe and Rugg [56].

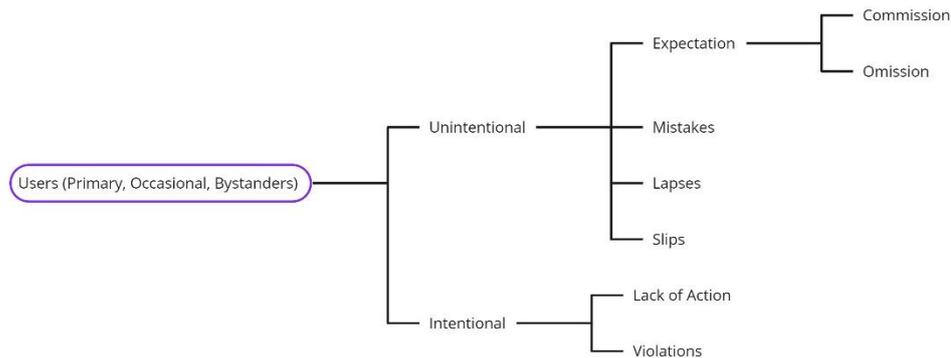

**Figure 6** Taxonomy of Failures caused by all types of Users.

*c.  Robot*

*Robot failures* (Figure 7) are divided into *Technical Errors* and *Social Errors*. Technical Failures (equivalent to system failures in Tolmeijer et a. [10]) refer to situations in which the system does not act as intended due to hardware errors (physical faults of the robotic equipment) or problems in the robot's software system. Social Failures (formally called 'social norm violations' in [2]) are errors that violate social norms and degrade a user's perception of a robot's socio-affective competence and their relationship with it [24]. Technical failures were quite detailed in [2], so only two amendments were made based on insights from the current study: (1) Design failures were moved from *Software failures* to *Service Provider,* to better represent the source of the issue; (2) "Body" was added under Hardware failures to represent structural deformities (e.g., dents). Social errors were broken down according to Tian & Oviatt [24].

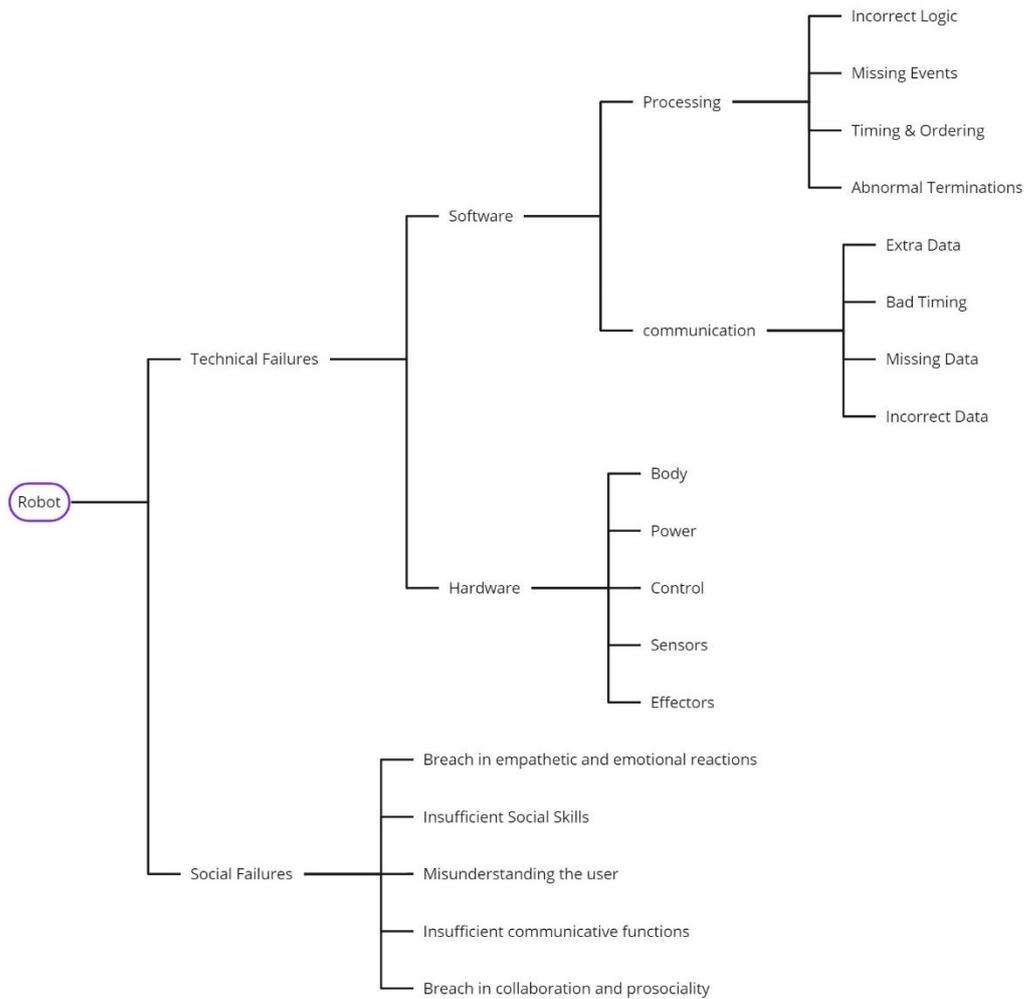

**Figure 7** Taxonomy of Failures caused by the Robot

*d. Relations between Sources*

The sources of failure described above are interconnected and influence one another. Their relations can be described as a funnel (Figure 8), where sources that are higher in the funnel influence the failures that occur from the sources below them. Failures introduced by the service provider could result in user errors that could cause the robot to fail. For example, the service provider of one of the robotic vacuum cleaners analyzed in the online reviews wanted their robot to identify and avoid cliffs (such as staircases) to prevent it from falling. Their cliff-identification algorithm was overly sensitive, resulting in robots confusing dark-colored carpets with cliffs (*design failure*). Whenever the robot would approach a dark carpet, it would report that it has encountered a drop and that it needed to be manually moved to another location. To resolve this issue, some customers covered relevant sensors with tape (*intentional user failure*). This solution, while effective for enabling the robot to clean dark carpets, often caused other issues with the robot (*robot* failures), such as the robot moving icoherently (*social failure*), running into obstacles, and falling down stairs (*technical failures*).

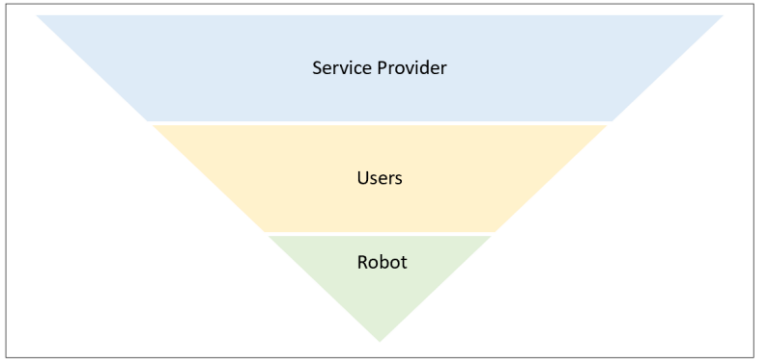

**Figure 8** Failure Funnel -  Sources that are higher in the funnel cause cascading failures in the sources below them.

## 9.     Discussion

One common challenge for robot developers and researchers is what problem to tackle first. For utilitarian domestic robots like those included in this analysis, technical failures related to Task Completion and Robustness & Resilience were the most commonly reported in online reviews (Figure 3). They had the most significant negative impact on customer experience (Figure 4), suggesting that failure-prevention and failure-response strategies should focus on the robot's technical ability to meet functional goals and maintain its operation and structural integrity over time. Our findings align with current development trends [2]. Prior research [10] attributed a higher risk score to software failures than hardware failures, stating that software failures are more common than hardware failures but that both are likely to have an equal impact on trust. Since we did not formally classify individual failures by hardware vs. software, we cannot compare their frequency or impact on customer experience. However, we can say that more identified failures in the reviews were attributed to software failures than hardware failures.

Yet, we have reasons to believe that hardware failures may be more impactful. Robustness & Resilience was the failure type that had the most significant negative impact on customer experience. The two most common complaints within this classification were that the robot stopped working or parts would break. These failures are more likely to be caused by hardware failures than software ones. Moreover, ease of repair was an occurring theme in the reviews that Tolmeijer et al.'s [10] risk analysis did not consider. Although possibly less common than software failures, hardware failures are more complex and expensive for customers to repair. They often require specialized tools, parts, and know-how that are inaccessible or costly to obtain. Many reviewers mentioned that their service provider did not have enough repair labs or technicians available for them locally, causing delayed repair times and the opportunity for problems to escalate. Some providers require customers to ship their robot at their own expense. Others tried to help customers (with varying success) identify the source of the problem remotely, then ask them to purchase and install new replacement parts on their own. This process is complicated, timely, and can introduce additional failures. In contrast, trained professionals can fix software failures remotely after deployment to the customer's home without much involvement from the user. For these reasons, we believe that efforts that prevent and help robots appropriately respond to hardware failures may be more impactful than software failures.

The failure types with the most negligible impact on star rating included Safety, Connection Issues & Updates, and User confusion (Figure 4), which may have resulted from their low representation within the dataset (Figure 3). The low impact of safety failures, situations in which the robot may hurt itself, someone else or lead to damage in the environment, on customer experience was particularly surprising, given the importance of safety to user acceptance in HRI [57, 58]. In addition to being rare, this finding may have been impacted by the fact that most of the tagged safety failures involved events of property damage or the potential thereof. Aside from a few anecdotes of the grill cleaning robots catching fire, most of these failures couldn't cause significant physical harm to humans, a tribute to the high safety standards set by the industry. In addition to producing genuinely safe robots, our results may also reflect that robotic companies successfully produce designs that promote perceptions of perceived safety to minimize the severity of potentially unsafe events. For example, one described problem was that the robots' wheels continued to turn even when lifted. Objectively, depending on the speeds at which robots continue to move and the material from which they're made, this could lead to significant injury, irrespective of the robot's size and wheels. Subjectively, this event may have been perceived as less of a threat because the robot's wheels were physically small. Perceived robot safety is a multifaceted construct that is impacted by many variables irrespective of objective safety, such as users' sense of control and comfort [59]. Further research is necessary to understand how accurate people are at assessing non-materialized robot threats to property or people and the extent to which events that are realized influence customer experience.

Issues related to usability and interaction design are essential but seem to be less detrimental to customer experience than technical failures. Although representing ~30% of all failures, interaction failures were found to impact customer experience the least. This can be seen both by their impact on star rating and their relatively high ratio of comments to complaints (Figure 3, Figure 4). We found evidence of all five social failure categories described in [24] in our dataset. User Effort, which we believe can be classified as a 'Breach in collaboration and prosociality' (specifically, a 'failure to maintain reciprocity or fairness in collaboration' [24]), is particularly interesting—it was one of the most commonly mentioned failures in the reviews, however, it did not have a large impact on customer experience. It is possible that this is a result of user expectations—that the majority of users expected robot operation to require effort on their part. Research has shown that forewarning users of potential failures can help mitigate their negative consequences [2, 60]. It is also possible that people don't mind putting in the effort if it is smaller than the effort required to have performed the task by other means; a sentiment expressed in several reviews. Previous studies have indicated that people have a general willingness to collaborate with robots and help them complete their tasks [61, 62]. However, it is unclear when and under which conditions willingness disappears. There seems to be a need to reduce user effort and improve fairness in collaboration, however, it does not seem to be as important to customers as getting the robot to be able to complete the job that it was "hired" to complete. The least common social failure category in our dataset was 'Breach in empathic and emotional reactions'. This may be the result of low user expectations of the robots' emotional abilities. Alternatively, it may be indicative of a general lack of interest in these types of social abilities for utilitarian robots. We think that as the robots' functional abilities improve, and user expectations adjust to better standards, social failures will become increasingly important to customers.

In addition to providing insights to inform failure prioritization efforts, the current investigation produced NLP models that could be used by designers and researchers of domestic robots to autonomously identify and monitor failure events of different levels of priority. The star ratings did not always represent our ability to extract meaningful information from a review—many 5-star reviews still had descriptions of failures in them, and many 1-star reviews were not descriptive enough (Table 3). This provides support for the usefulness of analyzing review content irrespective of star rating. Our models were able to predict with high confidence whether a given amazon review of a robotic vacuum cleaner described a failure or not (Table 6). To a lesser degree, they were also able to identify the type of failures it contains. Technical failures, that seem to be the most important to consumers, could be predicted with better accuracy and precision than interaction and service failures. Still, models of all three failure categories (technical, interaction, and service) performed significantly better than chance. Using NLP-based models to predict whether a given review contains a failure or not, and the type of failure it contains, could be used by researchers to estimate failure rates and compare them between robots with different attributes and users. Information regarding the extent to which a robot experiences failures is typically not readily available to researchers, since many companies prefer to prioritize perception of perfection over facilitating open communication and collaboration. Industry professionals could use these predictive models to conduct ongoing investigations of different types of robot failures, gain a better understanding of robot performance, and understand where to focus their development efforts.

By analyzing online customer reviews of a variety of domestic robots, we have found that the role that the service provider has on robot failures extends far beyond design failures in the robot itself. To be able to design robots that can appropriately respond to failures and/or help inexperienced users resolve them, it is important to consider the broader ecosystem within which human-robot interactions occur and interconnected relations between different sources of failure. It would be futile if a robot with tape placed on its sensor would respond to the issue by telling the user there is an obstruction on one of its sensors; of course there is, the user placed it there. It would be more helpful in this case if the robot and/or app first identified that this is a user violation and in response explained to the user the role that the sensor has in its general operation, possible consequences of it being obstructed, asked for the reason of violation, and provided a convenient way contact someone to help them overcome the issue (e.g. add a software feature that enables users to define a region in the room where the specific sensor will not be activated). This requires robots to be able to model chains-of-events of failures within the broader ecosystem, obtain information (either autonomously or by asking for help) regarding the likely source of failure, and adapt their responses accordingly. Future work should consider exploring how this could be done effectively. The importance of this and some initial efforts of how this could be done have been previously presented in the literature [10, 63], however, the current investigation produced concrete evidence from real-world encounters that highlight the extent of this need.

The present study has several limitations, which introduce opportunities for further research. Information can be gained only from what customers choose to report, and thus reflect a biased account of the human-robot interaction. The robotic failures that are being reported in online reviews are likely only a subset of the total failures that have been experienced in the real world. Not every person who noticed a problem with their robot will write a review, and not every person who writes a review will describe all the problems they have witnessed. There are a variety of factors that are likely to influence whether a person will describe a robotic failure within an online product review. First, the user must have noticed and comprehended that the failure had occurred in the first place. This ability depends on user characteristics, the symptoms of the failure, the robot's designed responses to such failures, and on the context in which the failure occurred. For an in-depth review of factors that influence people's ability to notice and comprehend robot failures, see [2]. Second, various personal characteristics impact the likelihood of a person to write a review. According to Pew Research Center, 38% of people never leave reviews on products and services and only about 10 percent "always" or "almost always" leave them. Studies have found that personal notions of attitude, subjective norm, emotional expression, ego-involvement, altruism, reciprocity, collectivism, enjoyment and self-efficacy were antecedent to writing online reviews [64–70]. Motivational differences were also found based on gender and income [65, 71]. Third, various contextual factors can influence the failures described in the review and how they are perceived by users, such as the amount of time the person spent interacting with the robot and type of interactions participants had with the robot before the issue took place, people's prior expectations regarding the robot's abilities, and the contents of recently written reviews. The above variability influences our ability to make definitive statements regarding the frequency of different types of failures and their impact on customer experience. Future research should try and collect or control for the above variables in order to gain a more accurate understanding of how often different types of failures occur, and how that influences their prioritization.

In addition, various scholars have argued that one-dimensional ratings, such as star-ratings, are unable to reflect the multiple aspects of customer opinions and the product experience [72, 73]. Many factors could have impacted the star-rating which are unrelated to the robots' failures, for example, self-efficacy [74], or the robot's pricing [20]. Future work assessing relations between failure types and customer experience should consider assessing customer experience in a multiplicated fashion; cross-referencing information obtained from a wide variety of sources, such as sentiment analysis [17], interviews and focus-groups. Another approach could involve measuring and accounting for confounding variables.

The most commonly reported issues we found in our study related to Task Completion, Robustness & Resilience and User Effort. This may indicate that robotic technologies are not yet reliably able to complete the tasks they were designed to accomplish without significant help from users. However, it may also simply be a reflection of the utilitarian nature of the dominant domestic robots on the market today. The types of failures commonly reported could be different for robots designed for other use cases, such as education and entertainment. Randall et al. [20] found "battery and charging issues" to be a common concern in product reviews on crowdfunding websites, whereas our analysis found that charging and battery issues were the least common. This discrepancy may be caused by the wider variety of robot applications included in their analysis. Another possible explanation is that since their analysis was based on the co-occurance of words, their findings do not indicate that battery and charging issues were more commonly reported, but rather that this is a problem that is shared by all robots, irrespective of their target application. The generalizability of our results to robots that are designed for different applications is unclear. Future studies should consider exploring the degree of similarity between robotic failures from different domains.

In the present research, we aimed to understand the types of failures users experience in domestic environments and how they influence customers' opinions. To do so, we analyzed thousands of online customer reviews of a variety of domestic robots and manually classified the failures people reported on in the reviews to identify potential sources and types of robot failure. To the best of our knowledge, this was the first and most extensive investigation into failed interactions between domestic robots and nonexpert users in natural environments. Through this investigation, we managed to identify sources and types of failures that had been previously overlooked in the literature. From a theoretical standpoint, we contributed by creating a new failure taxonomy that provides a broader perspective of how different failure sources and types relate to one another, while merging the state of the art on robot failures. From a practical standpoint, we contributed to the industry by developing and sharing novel NLP models that can be used to autonomously identify failures from the text of a review, and we provided actionable recommendations regarding the prioritization of design and development efforts based on the insights we collected. With this knowledge, designers and researchers of robotic systems can prioritize design and development efforts towards the most essential issues.

**Appendix:** All failures classified by type, category and their applicability to the different robot types included in the analysis. 1= failure was found in the reviews of the robot, R=Failure is relevant to this type of robot but no evidence of it was found in the reviews for the robot, N/A=this failure is not likely applicable for this type of robot.

| # | Failure | Type | Category | Vacuum Cleaner | Lawn Mower | Pool Cleaner | Grill Cleaner |
|---|---|---|---|---|---|---|---|
| 1 | Privacy concerns | Design / Service Quality | Service | 1 | 1 | R | R |
| 2 | Security concerns | Design / Service Quality | Service / Interaction | R | 1 | R | R |
| 3 | Issues with Pricing | Market Value | Service | 1 | 1 | 1 | 1 |
| 4 | Robot arrived in a used condition | Service Quality | Service | 1 | 1 | 1 | 1 |
| 5 | Issues with Customer Support or Customer Service | Service Quality | Service | 1 | 1 | 1 | 1 |
| 6 | Would like the base robot package to include more replacement parts | Service Quality | Service | 1 | 1 | 1 | R |
| 7 | Delivered without all items customer expected package to come with | Service Quality | Service | 1 | 1 | 1 | 1 |
| 8 | Social embarrassment | Service Quality | Service | 1 | R | R | 1 |
| 9 | Received wrong item | Service Quality | Service | 1 | 1 | 1 | 1 |
| 10 | Delivered broken or with loose parts | Service Quality | Service | 1 | R | 1 | 1 |
| 11 | Wanted more personalization choices | Service Quality | Service | 1 | R | R | 1 |
| 12 | Doesn't compare well with alternatives | Market Value | Service | 1 | 1 | 1 | 1 |
| 13 | General statements about the robot not meeting needs or expectations | Design | Service | 1 | 1 | 1 | 1 |
| 14 | Didn't like Colour | Design | Service | 1 | R | R | 1 |
| 15 | Issues with sizing of robot parts | Design | Service | 1 | 1 | 1 | R |
| 16 | the robot or the app is missing a specific function or ability the users want / user would like robot to be designed differently | Design | Service | 1 | 1 | 1 | 1 |
| 17 | Requires Wifi / App to access certain features or functions | Design | Service | 1 | R | 1 | R |
| 18 | App isn't supported by their devices | Design | Service | 1 | R | R | R |
| 19 | Robot or Feature isn't useful | Design | Service | 1 | 1 | 1 | R |
| 20 | Stopped working | Robustness & Resilience | Technical | 1 | 1 | 1 | 1 |

| # | Failure | Type | Category | Vacuum Cleaner | Lawn Mower | Pool Cleaner | Grill Cleaner |
|---|---|---|---|---|---|---|---|
| 21 | Parts breaking and wear & tear | Robustness & Resilience | Technical | 1 | 1 | 1 | 1 |
| 22 | inconsistent performance | Robustness & Resilience | Technical | 1 | 1 | 1 | 1 |
| 23 | Frequent Errors | Robustness & Resilience | Technical | 1 | 1 | 1 | R |
| 24 | Poor perceived quality | Robustness & Resilience | Interaction | 1 | 1 | 1 | 1 |
| 25 | Physical Button would not press or hard to press | Robustness & Resilience | Technical | R | R | 1 | 1 |
| 26 | Doesn't move in all directions | Robustness & Resilience | Technical | R | R | 1 | R |
| 27 | Robot isn't surfacing to top of pool | Robustness & Resilience | Technical | N/A | N/A | 1 | N/A |
| 28 | Brush bristles falling out on Grill | Robustness & Resilience | Technical | N/A | N/A | N/A | 1 |
| 29 | Remote doesn't respond well / work well / stopped working | Robustness & Resilience | Technical | 1 | R | 1 | R |
| 30 | App doesn't work well | Robustness & Resilience | Technical | 1 | R | R | 0 |
| 31 | Hard to repair | User Effort / Design | Service | 1 | R | 1 | R |
| 32 | Issues finding robot | User Effort | Interaction | 1 | 1 | N/A | N/A |
| 33 | Hard to clean robot/filter/bag | User Effort | Interaction | 1 | R | 1 | 1 |
| 34 | Hard to pick up and relocate | User Effort | Interaction | 1 | 1 | 1 | R |
| 35 | User Effort | User Effort | Interaction | 1 | 1 | 1 | 1 |
| 36 | The controls are hard to see | User Effort | Interaction | R | R | 1 | R |
| 37 | Takes a lot of time / effort to set up | User Effort | Interaction | 1 | 1 | 1 | R |
| 38 | Difficulties opening and closing / removing and adding components in robot | User Effort | Interaction | 1 | R | 1 | R |
| 39 | Didn't clean/mow everywhere | Task Completion | Technical | 1 | 1 | 1 | 1 |
| 40 | Doesn't pick up everything | Task Completion | Technical | 1 | 1 | 1 | 1 |
| 41 | Issues with how powerful suction/motor/cleaning ability is | Task Completion | Technical / Service | 1 | R | 1 | 1 |

| # | Failure | Type | Category | Vacuum Cleaner | Lawn Mower | Pool Cleaner | Grill Cleaner |
|---|---|---|---|---|---|---|---|
| 42 | would inexplicably cancel Job, turn itself off or have to be manually restarted | Task Completion | Technical | 1 | R | 1 | R |
| 43 | Spreads dirt | Task Completion | Technical | 1 | 1 | 1 | 1 |
| 44 | doesn't return to dock or takes a very long time to go back | Task Completion | Technical | 1 | 1 | N/A | N/A |
| 45 | gets stuck | Task Completion | Technical | 1 | 1 | 1 | 1 |
| 46 | Need to run it multiple times to complete the job | Task Completion / Task Completion Method | Technical | R | R | 1 | 1 |
| 47 | Brushes don't work well | Task Completion | Technical | R | N/A | 1 | 1 |
| 48 | Specialized filter doesn't function as well as the regular filter | Task Completion | Technical | R | N/A | 1 | R |
| 49 | Didn't seem like it was working | Task Completion | Interaction | 1 | R | R | 1 |
| 50 | General Statements about not cleaning / vacuuming / mowing well | Task Completion | Technical | 1 | R | 1 | 1 |
| 51 | Inaccurate | Task Completion | Technical | R | 1 | R | R |
| 52 | Robot gets clogged up from items in environment | Task Completion / User Effort / Incompatibility | Technical | R | N/A | 1 | N/A |
| 53 | The cable or power cord tangles or kinks | Task Completion / User Effort | Technical | N/A | N/A | 1 | N/A |
| 54 | The quick cycle which will tell the robot to not try and move up the walls doesn't work | Task Completion | Technical | N/A | N/A | 1 | N/A |
| 55 | Robot doesn't learn the space | Task Completion Method | Interaction | 1 | 1 | 1 | R |
| 56 | Robot repeatedly goes over the same area / runs around in circles | Task Completion Method | Interaction | 1 | 1 | 1 | R |
| 57 | Invests too little or too long in task | Task Completion Method | Interaction | 1 | 1 | 1 | 1 |
| 58 | Gets Lost | Task Completion Method | Technical / Interaction | 1 | 1 | N/A | N/A |
| 59 | Continues to go to problematic areas | Task Completion Method | Technical / Interaction | 1 | R | 1 | R |
| 60 | Not gratifying | Task Completion Method | Interaction | R | 1 | R | R |
| 61 | It doesn't always come out of the pool where I want it to | Task Completion Method | Interaction | N/A | N/A | 1 | N/A |

| # | Failure | Type | Category | Vacuum Cleaner | Lawn Mower | Pool Cleaner | Grill Cleaner |
|---|---|---|---|---|---|---|---|
| 62 | App is slow / Slow Response to Remote / Robot is slow | Task Completion Method | Technical / Interaction | 1 | 1 | 1 | N/A |
| 63 | Has difficulties in certain types of working environments | Incompatibility | Technical | 1 | 1 | 1 | 1 |
| 64 | Doesn't support all types of relevant working environments | Incompatibility / Design | Service | 1 | R | 1 | R |
| 65 | Pets don't get along with the Robot | Incompatibility | Interaction | 1 | 1 | R | R |
| 66 | Pets take out or move items necessary for robot operation | Incompatibility | Technical | 1 | 1 | R | N/A |
| 67 | Pet damaged robot | Incompatibility | Interaction | R | R | 1 | R |
| 68 | Scares people and animals in environment | Incompatibility | Interaction | 1 | R | R | 1 |
| 69 | Changes environment in a non-destructive yet undesirable way | Incompatibility | Interaction | 1 | 1 | 1 | N/A |
| 70 | Activates security alarm | Incompatibility | Interaction | 1 | R | N/A | N/A |
| 71 | Hair (human or pet) gets tangled on brush | Incompatibility | Interaction | 1 | R | R | 1 |
| 72 | Issues using the robot outside of the US | Incompatibility | Service | 1 | R | R | R |
| 73 | Has difficulties going over ledges | Incompatibility | Technical | 1 | R | N/A | N/A |
| 74 | loud noises in inappropriate times | Incompatibility | Interaction | 1 | R | R | R |
| 75 | Too loud / noisy | Incompatibility | Interaction | 1 | R | 1 | 1 |
| 76 | Cannot handle water / it's not waterproof / water damaged robot | Incompatibility | Technical / Service | 1 | R | 1 | R |
| 77 | issues working alongside other devices in the environment | Incompatibility | Interaction | R | R | 1 | N/A |
| 78 | Makes the grill shake / rattle | Incompatibility | Interaction | N/A | N/A | N/A | 1 |
| 79 | The robot gets very dirty | Incompatibility | Interaction | R | R | R | 1 |
| 80 | robot alerted for a heat that was lower than what it should be able to handle | Incompatibility / User Confusion / Design | Service | N/A | N/A | N/A | 1 |
| 81 | You can't swim when it's in the pool | Incompatibility / User Effort | Interaction | N/A | N/A | 1 | N/A |
| 82 | Issues with storage | Incompatibility / Design | Service | R | 1 | 1 | 1 |
| 83 | Robot is damaged by interacting with operating environment | Incompatibility / Safety / Robustness & Resilience | Technical | R | R | 1 | R |
| 84 | Runs into walls and obstacles | Safety | Technical | 1 | 1 | 1 | 1 |
| 85 | Robot moves obstacles that it runs into | Safety | Technical | 1 | 1 | R | N/A |

| # | Failure | Type | Category | Vacuum Cleaner | Lawn Mower | Pool Cleaner | Grill Cleaner |
|---|---|---|---|---|---|---|---|
| 86 | Damaged floor or items in the environment | Safety | Technical | 1 | 1 | 1 | 1 |
| 87 | Wanders where it shouldn't | Safety | Technical | 1 | 1 | R | 1 |
| 88 | The robot and wheels don't stop moving when picking up the robot / taken out of the pool | Safety | Technical | 1 | R | 1 | R |
| 89 | Caused person or animal pain / harm / ran into animal / person | Safety | Technical | 1 | R | 1 | 1 |
| 90 | Robot eats items | Safety | Technical | 1 | 1 | R | R |
| 91 | Concerns about potential safety issues | Safety | Interaction / Service | R | 1 | R | 1 |
| 92 | Caused Fire or Smoke | Incompatibility / Safety / Robustness & Resilience | Technical | R | R | 1 | 1 |
| 93 | Misunderstood robot operating requirements or abilities | User Confusion | Service / Interaction | 1 | 1 | 1 | 1 |
| 94 | Incompatibility between robot feedback and task/behavior robot is performing | User Confusion | Interaction | 1 | R | 1 | 1 |
| 95 | Unclear indicators | User Confusion | Interaction | 1 | R | 1 | 1 |
| 96 | Technical Issues with scheduling | User Confusion / Connection Issues & Updates | Technical | 1 | R | R | N/A |
| 97 | Confusing, hard to learn, use or set up | User Confusion | Interaction | 1 | 1 | 1 | R |
| 98 | Trouble downloading the app | User Confusion | Interaction | 1 | R | R | R |
| 99 | Doesn't come with manual or instructions for how to operate or fix robot | User Confusion | Interaction | 1 | R | 1 | 1 |
| 100 | Delayed notifications | User Confusion | Interaction | 1 | R | R | R |
| 101 | unclear or incorrect operating instructions | User Confusion | Interaction | 1 | 1 | 1 | 1 |
| 102 | missing indicators | User Confusion | Interaction | R | R | 1 | 1 |
| 103 | Unsure about the state of the robot | User Confusion | Interaction | R | R | 1 | R |
| 104 | Usability issues while registering robot | User Confusion | Interaction | R | R | 1 | R |
| 105 | Produced erroneous sounds | User Confusion / Robustness & Resilience | Technical | 1 | R | 1 | R |
| 106 | Issues connecting to charger / power supply | Charging & Battery | Technical | 1 | 1 | 1 | |

| # | Failure | Type | Category | Vacuum Cleaner | Lawn Mower | Pool Cleaner | Grill Cleaner |
|---|---|---|---|---|---|---|---|
| 107 | Battery charge doesn't last long enough | Charging & Battery / Robustness & Resilience | Technical / Interaction | 1 | 1 | 1 | 1 |
| 108 | Battery would not charge / defective battery or power cord | Charging & Battery / Robustness & Resilience | Technical | 1 | 1 | R | 1 |
| 109 | Slow charging | Charging & Battery | Technical / Interaction | 1 | R | R | R |
| 110 | Power cord is too short | Charging & Battery | Interaction / Service | R | R | 1 | R |
| 111 | Battery came depleted | Charging & Battery / Service Quality | Service / Technical | R | R | R | 1 |
| 112 | Unit turns itself off after a few moments | Charging & Battery / Robustness & Resilience | Technical | R | R | 1 | R |
| 113 | If there is not enough sun it will run out of solar power quite quickly | Charging & Battery / Incompatibility | Interaction | N/A | R | 1 | R |
| 114 | Problems connecting Google voice control / Alexa / App to control the robot | Connection Issues & Updates | Interaction / Technical | 1 | R | R | R |
| 115 | Fails to connect to Wifi / disconnects from Wifi / app doesn't connect to Robot / | Connection Issues & Updates | Technical | 1 | 1 | 1 | R |
| 116 | wasn't responding to programming | Connection Issues & Updates / Task Completion | Technical | 1 | R | 1 | R |
| 117 | Infrequent updates and upgrades | Connection Issues & Updates | Service | 1 | R | R | R |
| 118 | Problems completing software update | Connection Issues & Updates | Interaction / Technical | 1 | R | R | 1 |
| 119 | If robot got clogged and you are not around, you will end up wasting electricity for the remaining duration that you set it up initially for. | Task Completion / Charging & Battery | Interaction | N/A | N/A | 1 | N/A |